\documentclass{article} 
\usepackage{iclr2025_conference,times}

\usepackage{amsmath,amsfonts,bm}

\def\eqref#1{equation~\ref{#1}}

\def\1{\bm{1}}

\DeclareMathAlphabet{\mathsfit}{\encodingdefault}{\sfdefault}{m}{sl}
\SetMathAlphabet{\mathsfit}{bold}{\encodingdefault}{\sfdefault}{bx}{n}

\DeclareMathOperator*{\argmax}{arg\,max}

\usepackage{hyperref}
\usepackage{url}
\usepackage{listings}
\usepackage{hyperref}
\usepackage{mathrsfs}
\usepackage{xfrac}    
\usepackage[most]{tcolorbox}
\usepackage{amssymb}
\usepackage{color}
\usepackage{graphicx}
\usepackage{booktabs}
\usepackage{enumitem}
\usepackage{forest}
\definecolor{keywordcolor}{rgb}{0.7, 0.1, 0.1}   
\definecolor{tacticcolor}{rgb}{0.0, 0.1, 0.6}    
\definecolor{commentcolor}{rgb}{0.4, 0.4, 0.4}   
\definecolor{symbolcolor}{rgb}{0.0, 0.1, 0.6}    
\definecolor{sortcolor}{rgb}{0.1, 0.5, 0.1}      
\definecolor{attributecolor}{rgb}{0.7, 0.1, 0.1} 
\def\lstlanguagefiles{lstlean.tex}

\title{\textsc{HunyuanProver}: A Scalable Data Synthesis Framework and Guided Tree Search for Automated Theorem Proving}

\author{Yang Li\thanks{Equal contribution}, Dong Du*, Linfeng Song*, Chen Li*, Weikang Wang, Tao Yang and Haitao Mi \\
Tencent Hunyuan Teams \\
\texttt{\{youngyli,dongdu,vegetali,daxianwang,rigorosyang\}@tencent.com} \\
\texttt{\{lfsong,haitaomi\}@global.tencent.com} \\
}

\iclrfinalcopy 
\begin{document}

\maketitle

\begin{abstract}
We introduce \textsc{HunyuanProver}, an language model finetuned from the \textsc{Hunyuan 7B} for interactive automatic theorem proving with LEAN4.
To alleviate the data sparsity issue, we design a scalable framework to iterative synthesize data with low cost.
Besides, guided tree search algorithms are designed to enable effective ``system 2 thinking`` of the prover.
\textsc{HunyuanProver} achieves
state-of-the-art (SOTA) performances on major benchmarks.
Specifically, it achieves a pass of 68.4\% on the miniF2F-test compared to 65.9\%, the current SOTA results.
It proves 4 IMO statements (\textsc{imo\_1960\_p2}, \textsc{imo\_1962\_p2}, \textsc{imo\_1964\_p2} and \textsc{imo\_1983\_p6}) in miniF2F-test.

\end{abstract}

\section{Introduction}

Recent advancements in large language models (LLMs) have profoundly impacted mathematical reasoning and theorem proving in artificial intelligence.
Despite notable progress in natural language domains, language models still encounter substantial challenges in formal theorem proving (e.g., using LEAN \citep{moura2021lean} or Isabelle \citep{paulson1994isabelle}) probably due to the massive search space of Olympiad-level theorem proving and 
limited data in such scenario.
As the results, even the most advanced models like GPT-4o \cite{hurst2024gpt} struggle with complex formal-statement proving \citep{xin2024deepseekv1}.
A formal theorem-proving model should be capable of understanding both the syntax and semantics of a formal system, enabling it to generate valid next steps within the system. More critically, it must integrate this capability with its abstract mathematical reasoning skills to perform effective and efficient deductions.

We propose \textsc{HunyuanProver}, an LLM-based automatic theorem prover obtained by a pragmatic framework for addressing the aforementioned challenges.
The framework takes two core modules: a scalable prover-data generator and guided tree-search algorithms.
The prover-data generator only takes open-source data for theorem proving to train the initial autoformalizer and prover.
The autoformalizer then converts a large amount of existing math questions into the format of the target prover (e.g., LEAN4).
The prover is lastly iteratively improved on such data where new proof data is generated at each iteration to train the prover.
At testing time, tree-search algorithms and multiple critic models are designed to perform ``slow thinking'' that is empirically essential for solving complex theorem-proving tasks.

Evaluations show that \textsc{HunyuanProver} yields an accuracy of 68.4\% on the miniF2F benchmark.
In addition, we also observe several key findings: 1) using explicitly trained critic for tree-search guidance is helpful; 2) the scale of finetuning data for theorem proving is critical, thus designing efficient data scaling framework is important; 3) data curation and selection is important as well when there is sufficient amount of training data.

Our key contributions includes:
\begin{itemize}[leftmargin=*]
    \item We introduce a scalable pipeline that utilizes open-source data for theorem proving and massive math problems in natural language to generate a large number of new training data for automatic theorem proving.
    \item We develop several critic models and evaluate their effectiveness using two widely adopted search algorithms: best-first search and Monte Carlo tree search. 
   
\end{itemize}

\section{Scalable Data Generation for Prover Improving}
\label{sec:policy_improve}

One major bottleneck for automated theorem proving is the lack of training data.
For example, as one of the largest open-source LEAN4 datasets, mathlib4 \citep{moura2021lean} only contains around 50k theorems(with tactics) for training.
This is far from sufficient to train a stronger prover given the difficulty of automatic theorem proving.
On the other hand, massive high-quality math problems in natural language have been released in recent years \citep{yumetamath,luo2023wizardmath}.

Our efforts fall into two critical aspects to scale the training data for automatic theorem proving: autoformalization (Section \ref{sec:form_data}) that maps a natural language problem (Section \ref{sec:step_data}) into LEAN format statement, and tactic generation for iterative theorem proving.

\subsection{Autoformalization Data Generation}
\label{sec:form_data}

For autoformalization, we start with 130k high-quality natural language to LEAN statement pairs, which includes 50k from lean workbook \citep{ying2024lean} and 80k from MMA \citep{jiang2024multilingual}.
We first translate the natural language part of the 130k data into Chinese to double the dataset size, and with such data we train a autoformalization model, it can translate both English and Chinese problems to lean4 format statement.

In the next step, 30 million internal math problems in natural language are converted into formal statement by the autoformalization model. For each natural language problem, we sample 8 outputs with different temperatures.We obtain a dataset $D^q$ of 20 million LEAN statements after filtering out these do not conform with LEAN grammar or do not satisfy other rules adopted by previous practices \citep{ying2024lean}.

In addition to using internal data, we also utilized open-source natural language mathematical data NuminaMath CoT\citep{numina_math_datasets}.
\subsection{Prover Improving via Iterative Tactic-Level Proving Data Generation}
\label{sec:step_data}

We design a iterative framework that takes a LEAN engine $\mathbf{\Gamma}$ and statement dataset $D^q$ from autoformalization to generate new tactic data for prover improving.
Specifically, in iteration $t$, we leverage best-first search (Section \ref{sec:tsa}) algorithm with the prover from the previous iteration $\pi_{t-1}$ on all unsolved statements so far in $D$.
We collect the proving trajectories (e.g., $\tau$) for newly solved statements (e.g., $q$) if there is any:
\begin{equation}
    D_t = \{(q,\tau)~|~q\in D^q-D_{t-1}, \tau\sim\textsc{Bfs}(q),\tau \ne null\} \cup D_{t-1}
\end{equation}
Then the prover is updated using rejection finetuning (RFT) with the proving trajectories in $D_t$ after filtering out easy statements solved in early iterations.
The initial prover $\pi_0$ is trained on public data including mathlib4.
After more than 10 rounds of iteration, more than 20 million tactic-level data is obtained.

\paragraph{Enhancing Diversity}
As indicated by previous work \citep{xin2024deepseek}, prover diversity is important due to the massive search space of theorem proving.
We further develop two methods to enhance the prover diversity within the iterative data-generation framework.
For the first method, we design rules to convert the last state of an unfinished proving trajectory into a new statement.
In this way, more diverse proving data is obtained for prover training.
For the second method, we collect data from proving more challenging statements, including those Olympiad-level algebraic inequalities \cite{wei2024proving} and lean workbook \cite{ying2024lean}.

\begin{figure}
    \centering
    \includegraphics[width=0.95\linewidth]{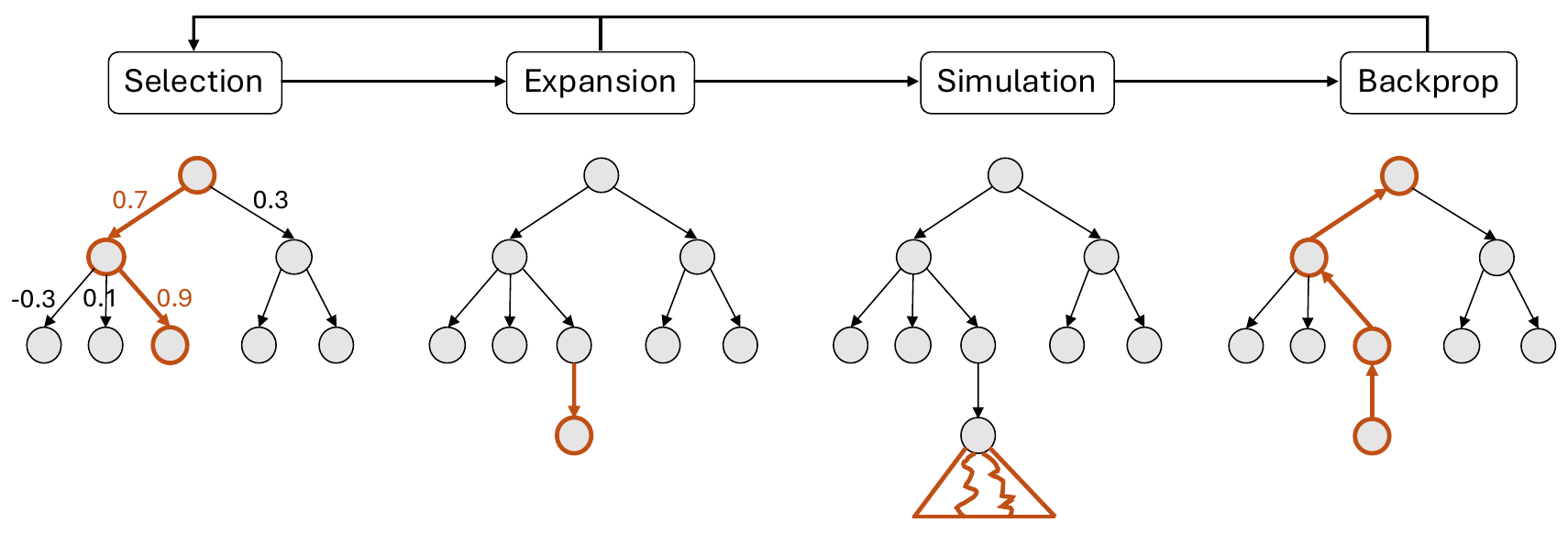}
    \caption{Comparing best-first search (BFS) with Monte-Carlo tree search (MCTS). BFS only takes \emph{Selection} and \emph{Expansion} in one iteration, while MCTS takes all four steps. The numbers represent critic-assigned scores.}
    \label{fig:tree_search_demo}
\end{figure}

\section{Guided Tree Search}
\label{sec:tree_search}

As described in Section \ref{sec:policy_improve}, our task involves iterative interaction with LEAN as the environment, where for each iteration the policy predicts a new tactic given from a state in the proving process.
We abstract this process as tree search where a state $s_i$ in the proving process corresponds to a tree node $n_i$ with the input statement $q$ being the root $n_0$.
An edge that points from $n_i$ to $n_j$ represents applying a tactic on node (state) $n_i$ to yield node (state) $n_j$.
To handle this problem, we design two major tree search algorithms as described in Section \ref{sec:tsa}.
We also design several critics, as shown in Section \ref{sec:critic}, to guide these algorithms.

\subsection{Tree Search Algorithms}
\label{sec:tsa}

We explore two search algorithms: best-first search (BFS) and Monte-Carlo tree search (MCTS) for the tactic-level iterative statement proving process.

\paragraph{Best-First Search}
We first study BFS due to its simplicity and effectiveness.
As demoed in the left part of Figure \ref{fig:tree_search_demo}, BFS can be viewed as an iterative process of \emph{selection} and \emph{expansion}.
The search process continues until either the statement $q$ is successfully proved or the maximal number of iteration $T$ is reached.

In the selection step, the node $\hat{n}$ with the highest critic score is selected (without return) from the set of active nodes $\mathcal{N}$:
\begin{equation}
    \hat{n} = \argmax_{n\in \mathcal{N}} \textsc{Critic}(n)
\end{equation}
where $\textsc{Critic}$ represents the critic function.
In the expansion step, the policy $\pi$ samples a set ($\hat{\mathcal{C}}$) of $K$ candidate tactics under $\hat{n}$:
\begin{equation}
    \hat{\mathcal{C}} = \{\hat{c}_i | \hat{c}_i \sim \pi (q, \hat{n}), i\in [0,\dots,K]\}
\end{equation}
Each tactic in $\hat{\mathcal{C}}$ is then executed against the \textsc{LEAN} engine to yield a new tree node if the tactic is valid under state $\hat{n}$.
After removing those that are identical to previously explored tree nodes, the remaining are merged into the active node set $\mathcal{N}$.
We check two nodes are identical simply by string matching.

\paragraph{$\eta$MCTS}
Thought being simple and effective in general, BFS has several limitations on handling complex search problems.
One limitation is that each node is only visited once with a fixed amount of expansion budget.
Besides, BFS only takes the critic score as guidance, thus it can suffer from any bias and misjudgment inherent in the critic model.

We additionally adapt $\eta$MCTS \citep{tian2024toward} to handle such limitations. 
As shown in Figure \ref{fig:tree_search_demo}, the original $\eta$MCTS algorithm takes 4 steps of \emph{selection}, \emph{expansion}, \emph{simulation} and \emph{back-propagation} in each iteration.
Here we remove the step of \emph{simulation}, leaving it for future work.
As another major, our $\eta$MCTS follows the setting of BFS to sample $K$ candidate tactics (instead of one in \cite{tian2024toward}) at a time given a state.
Different from BFS, our $\eta$MCTS algorithm can sample a tree node multiple times, and the expansion budget for each node is updated by its importance score that can be dynamically changed throughout the search process.
Following \cite{tian2024toward}, the importance score for any node $n$ is defined as the maximal value difference between it and its descendant:
\begin{equation}
    I(n) = \max_{\hat{n}\in\textsc{Succ}(n)}|\textsc{Critic}(\hat{n}) - \textsc{Critic}(n)|,
\end{equation}
where $\textsc{Succ}(n)$ represents all succeeding nodes of node $n$.
Then, the adapted expansion budget for node $n$ is defined as:
\begin{equation}
    E(n) = \max(B_{min}, \min(B_{max}, \lfloor \alpha I(n) \rfloor + 1))
\end{equation}
where $\alpha$, $B_{min}$, $B_{max}$ are the corresponding factor, lower bound and upper bound, respectively.

As another major difference from BFS, $\eta$MCTS selects nodes based on Upper Confidence Bound (UCB) defined as:
\begin{equation}
    UCB(n) = \textsc{Critic}(n) + \alpha \times \sqrt{2\times \ln \frac{\textsc{Cnt}(\textsc{Prnt}(n))}{\textsc{Cnt}(n)}}
\end{equation}
where $\textsc{Prnt}(n)$ denotes the parent node of $n$ in the search tree, and $\textsc{Cnt}(n)$ represents the number of visiting times so far for $n$.
Generally, UCB can be viewed as a balance between exploration and exploitation, where the first term of the equation prefers nodes with high critic score (\emph{exploitation}) while the second term prefers under-explored nodes (\emph{exploration}).

\subsection{Critics for Search Guidance}
\label{sec:critic}

Critic modeling is a central component in tree search as it provides the guidance.
We design three types of critic models to guide the search process.

\paragraph{Policy Confidence (PC)}
We first leverage policy confidence as guidance for a cold start of guided search due to limited tree-search data for training critic models at the beginning.
Particularly, for tactic $c$ under state $n$ in the proving process of statement $q$, we define its policy confidence $f^\pi(c)$ as token-level average log probability:
\begin{equation}
    f^\pi(c) = \frac{1}{|c|}\sum_{j=1}^{|c|} \log p_\pi (c^j | q, n, c^{<j})
\end{equation}
where $|c|$ is the number of tokens in $c$.

\paragraph{Process Reward Model (PRM)}
The process reward model (PRM), denoted as $v_\phi^\pi(q, n)$ (parameterized by $\phi$), represents the possibility of tree node $n$ for proving statement $q$ when starting from it and following policy $\pi$ thereafter. 

To train a parameterized PRM $v_\phi^\pi$ using statement set $D = [q^1,\dots]$, we first generate a search tree for each statement $q^k$ by following policy $\pi$ under the guidance of the critic from the previous iteration.
Next, for each node $n^k_i$ in the search tree, a score is assigned to reflect its quality.
Ideally, human experts are hired for rating the node quality, but the cost is dramatic.
Following \cite{wang2024math}, we simply assign
a score of +1 or -1 to each node by
indicating whether it can reach the final state that indicates successful proving of $q^k$.
This approximation has been proven effective in the experiments of \cite{wang2024math}.
As the results, we obtain a PRM dataset $D_{v} = [(q^k, n^k_i, l^k_i),\dots]$ with $l^k_i$ indicating the score for node $n^k_i$.

With dataset $D_{v}$, the PRM is then optimized by minimizing the mean squared error for each node:
\begin{equation}
v_\phi^\pi = -\mathbb{E}_{(q^k, n^k_i, l^k_i)\sim D_{v}}(v_\phi^\pi(q^k, n^k_i) - l^k_i)^2
\end{equation}
Similar to \cite{wang2024math}, $v_\phi^\pi$ is a LLM with an MLP layer on top to output a scalar for each token.
We use the scalar prediction at the last token of each state as the value.

\paragraph{Distance Critic (DC)}
PRM only captures the possibility of proving the input statement $q$ starting from a tree node $n$,
therefore we further design a distance-based critic model to predict the estimated remaining number of steps to prove $q$ from $n$.

One naive solution can be training the critic model to directly predict the distance, while it is likely to suffer from data sparsity issue. When the state is very complex, it is very difficult to predict an accurate number of the remaining steps. 
We train the critic model to predict not only the exact number of remaining steps but also the range in which this number falls, represented using a balanced binary tree structure.
This approach helps mitigate the data sparsity issue by enabling the model to make predictions in a hierarchical, coarse-to-fine manner. Specifically, the critic model is trained to identify a path on the balanced binary tree, progressively narrowing down to the exact number of remaining steps.

\begin{figure}[h]
\centering
\begin{forest}
[root,red [1/2 [1/4 [1/8] [2/8]] [2/4 [3/8] [4/8] ]][2/2,red,edge=red [3/4,red,edge=red [5/8] [6/8,red,edge=red]][4/4 [7/8][8/8]]]]
\end{forest}
\caption{Balanced binary tree structure for coarse-to-fine number representation. When the level of tree is 4, the max number can be represented is 8. The path of number $6$ is $root \rightarrow 2/2 \rightarrow 3/4 \rightarrow 6/8$. The tuple associated with the path is (2, 3, 6).}
 \label{fig:path_based_dc}
\end{figure}
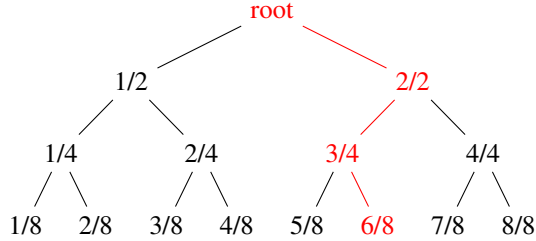

Figure \ref{fig:path_based_dc} illustrates a 4-level binary tree capable of representing numbers from 1 to 8. Each level of the tree corresponds to a progressively finer division of the numerical range. For instance, at the second level, there are two nodes: $1/2$ and $2/2$, which split the entire range into a left half $[1,4]$ and a right half $[5,8]$. If a number falls within the range $[1,4]$, the corresponding path through the tree includes the node 
$1/2$.
Empirically, nodes at higher levels of the tree are easier to predict than leaf nodes. This hierarchical structure helps reduce prediction errors during tree search by enabling the model to first make broader, more reliable predictions before refining them.
For example, two numbers that are close in distance share a common prefix in their paths on the tree. Therefore, representing numbers using their paths on the tree effectively captures the distance relationship between them, enabling the model to predict this information more directly and intuitively.

In practice, we use an 8-level binary tree to represent numbers up to 64, where each node on the tree is represented by a separate special token. For instance, 1/2 is represented by the token \texttt{\textless\textbar num-1-of-2\textbar\textgreater}, and 5/8 is represented by \texttt{\textless\textbar num-5-of-8\textbar\textgreater}.
When constructing training data, if the number of remaining steps exceeds 64, we set it to 64.

A path on the binary tree is essentially a tuple of numbers, such as (2,3,6) shown in Figure \ref{fig:path_based_dc}. During the tree-search stage with the distance critic, we compare two states by directly comparing their corresponding tuples. This approach inherently evaluates states in a coarse-to-fine manner.




\section{Experiments}

\subsection{Setup}

\paragraph{Benchmarks}
We evaluate theorem-proving performance on the following benchmarks to pinpoint the effectiveness of each proposed module:
\begin{itemize}[leftmargin=*]
    \item MiniF2F \citep{zheng2022miniff} examines LLM's automatic formal problem-solving skills targeted at high-school-level exercises and competitions, such as AMC, AIME, and IMO, with a particular focus on algebra and number theory. The benchmark comprises 244 validation problems and 244 test problems.
\end{itemize}

\paragraph{Inference}
We use the LEAN engine from LeanDojo \citep{yang2024leandojo} as the engine to conduct both tactic data generation and benchmark evaluation.
Whole and per-step timeout for tactic executing in LEAN are set to 3600 seconds and 60 seconds, respectively.
At most 800 search steps are conducted for both BFS and MCTS. 
For each search step, $K=8$ tactics are sampled from the given LEAN state under temperatures 0.7, 0.8, 1.0 and 1.1, where two tactic are sampled under each temperature.
These temperature values are empirically decided.
The prompts for policy and distance critic are listed in Appendix \ref{sec:policy_prompt} and \ref{sec:distance_critic_prompt}.

\paragraph{Finetuning Hyperparameters}
Our prover is obtained by fine-tuning a \textsc{Hunyuan 7B} model on our self-generated tactic data.
During fine-tuning, at most 4 epochs are conducted and checkpoint is selected based on miniF2F valid set.
The maximum sequence length, learning rate, minimal learning rate and batch size are set to 4096, $2\times 10^{-5}$, $1\times 10^{-6}$ and 256, respectively.
Cosine learning schedule is used. 

\paragraph{Comparing Systems}
We compare HunyuanProver with former state-of-the-art systems.
Among them, Lean-STaR \citep{lin24lean} and InternLM2.5-StepProver \citep{wu2024internlm2} are interactive step-level proving methods, 
while on the other hand, DeepSeek-Prover-V1.5 \citep{xin2024deepseekproverv15} is a whole-proof generation method.

\subsection{Results and Analysis}

\begin{table}[]
    \centering
    \begin{tabular}{lcccc}
    \toprule
    System & Model Size & Sample Budget & MiniF2F-test \\
    \midrule
    \emph{Whole-Proof Generation Methods} \\
    DeepSeek-Prover-V1.5-RL+MCTS & 7B & 16 × 6400 & 60.2\% \\
    DeepSeek-Prover-V1.5-RL+RMaxTS & 7B & 32 × 6400 & 63.5\% \\
    \midrule
    \emph{Interactive Step Proving Methods} \\
    Lean-STaR+BFS+CG & 7B & 64 × 1 × 50 & 46.3\% \\
    InternLM2.5-StepProver+BFS & 7B & 256 × 32 × 600 & 59.4\% \\
    InternLM2.5-StepProver+BFS+CG & 7B & 256 × 32 × 600 & 65.9\% \\
    HunyuanProver v16+BFS & 7B & 600 × 8 × 400 & 64.8\% \\
    HunyuanProver v16+BFS+DC & 7B & 600 × 8 × 400 & \textbf{68.4}\% \\
    \bottomrule
    \end{tabular}
    \caption{Main comparison regarding accuracy and sampling cost with other systems on MiniF2F-test. CG indicates critic-guided search, while DC represents taking our proposed distance critic as guidance. For BFS methods, the cost is represented as \#Pass × \#Beam × \#Iteration, while for MCTS, the cost is defined as \#Pass × \#Iteration.}
    \label{tab:main}
\end{table}

Table \ref{tab:main} shows the comparison between HunyuanProver with other existing state-of-the-art systems.
Among these systems, our HunyuanProver with distance-tritic as guidance shows the best accuracy on miniF2F, advancing the previous SOTA system (InternLM2.5-StepProver+BFS+CG) by 2.5\% points using less search budget.
Though InternLM2.5-StepProver+BFS+CG is slightly better than DeepSeek-Prover-V1.5-RL+MCTS, the later does not take any critic for guidance.
We believe MCTS can be more beneficial to the task of automatic theorem proving, and this potential has not been fully revealed.
In addition, we can also probe the importance of critic guidance by incorporating DCG on HunyuanProver, which shows 3.6\%-point gain.
We list multiple statements solved by HunyuanProver in Appendix \ref{sec:examples}.

\paragraph{Effectiveness of Iterative Tactic Data Generation}

\begin{figure}[t]
    \centering
    \includegraphics[width=0.9\linewidth]{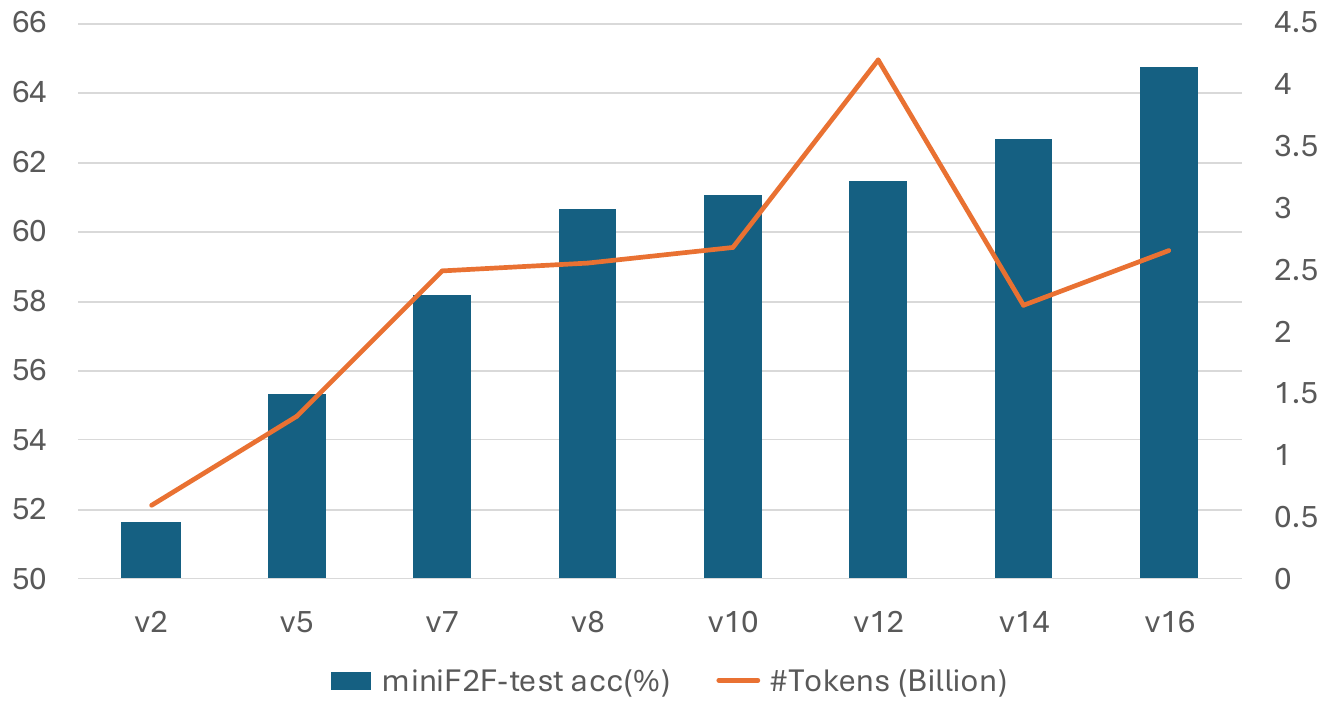}
    \caption{The trend regarding miniF2F-test accuracy and the number of finetuning tokens during the iterative tactic data generation process for prover improving, where ``v'' represents ``version''. The version number is approximately equivalent to the number of iterations. After v12, we remove some easy training data. BFS with policy confidence as the critic is adopted.}
    \label{fig:iter_data_gen}
\end{figure}

Figure \ref{fig:iter_data_gen} visualizes the changes regarding miniF2F-test accuracy and the amount of training data during the prover improving process based on iterative tactic data generation.
We initially see performance boosts in early iterations until version v8, and then minor improvements are observed by further increase the number of training tokens from roughly 2.75B (v8) to 4.25B (v12).
After v12, we remove some data generated from early iterations(before v8), Most of the removed training data are relatively easy statements.We can see the performance boost is achieved by removing these easy data.
This highlights the importance of data selection in the iterative improving process.

\paragraph{Effectiveness of Different Critics and Tree Search Methods}
As shown in Table \ref{tab:ablation_gts}, we conduct an ablation study on different search and critic combinations under three different versions of HunyuanProver.
Due to limitation on time and computation resources, MCTS is executed together with PRM.
The MCTS-with-PRM combination is consistently better than the BFS with policy confidence.
Due to the limitation of time and computation resources, we leave separately examining the effectiveness of MCTS and PRM in future work.
In addition, simply replacing policy confidence with distance critic is also significant effective.
Since both PRM and DC only require prover generated data with natural labels, thus they are both superior choices than policy confidence as critic.

\begin{table}[]
    \centering
    \begin{tabular}{lc}
    \toprule
    Model + Search & MiniF2F-test \\
    \midrule
    \textsc{HunyuanProver v12} \\
    ~~~+ (BFS, PC) & 61.07\% \\
    ~~~+ (MCTS, PRM) & 62.29\% \\
    \midrule
    \textsc{HunyuanProver v14} \\
    ~~~+ (BFS, PC) & 62.70\% \\
    ~~~+ (BFS, DC) & 65.57\% \\
    ~~~+ (MCTS, PRM) & 66.39\% \\
    \midrule
    \textsc{HunyuanProver v16} \\
    ~~~+ (BFS, PC) & 64.75\% \\
    ~~~+ (BFS, DC) & 68.44\% \\
    \bottomrule
    \end{tabular}
    \caption{Ablation study on different variations of guided tree search.}
    \label{tab:ablation_gts}
\end{table}

\begin{figure}[t]
    \centering
    \includegraphics[width=0.9\linewidth]{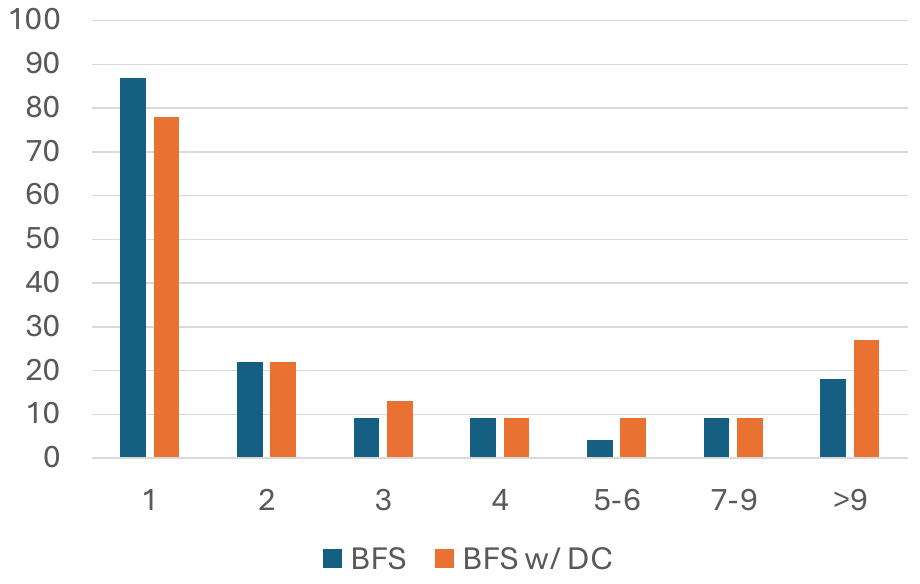}
    \caption{Proof length distributions of BFS using policy confidence and distance critic respectively for guidance.}
    \label{fig:minif2f_dist}
\end{figure}

Figure \ref{fig:minif2f_dist} visualizes the distribution over solved miniF2F-test theorems for HunyuanProver v16 and HunyuanProver v16+DC.
HunyuanProver v16+DC finds more deep proofs compared to HunyuanProver v16 in miniF2F.
Interestingly, HunyuanProver v16+DC takes more steps than HunyuanProver v16 on some of the easiest theorems.
We follow previous work to calculate the proof length based on the number of steps in the shortest proof for each theorem.

\section{Conclusions}

In this report, we present HunyuanProver, a system that enhances automatic theorem-proving capabilities using LEAN through iterative tactic data generation and guided tree search. The iterative tactic data generation method expands the training dataset by nearly 40-fold, resulting in a significant performance improvement. Meanwhile, critic-guided tree search algorithms further enhance the overall effectiveness of the prover.

Future work includes better curation for prover training data and exploring other more cost-efficient tree search algorithms such as Q* \citep{wang2024litesearch,wang2024q}.


\bibliography{iclr2025_conference}

\begin{thebibliography}{20}
\providecommand{\natexlab}[1]{#1}
\providecommand{\url}[1]{\texttt{#1}}
\expandafter\ifx\csname urlstyle\endcsname\relax
  \providecommand{\doi}[1]{doi: #1}\else
  \providecommand{\doi}{doi: \begingroup \urlstyle{rm}\Url}\fi

\bibitem[Hurst et~al.(2024)Hurst, Lerer, Goucher, Perelman, Ramesh, Clark, Ostrow, Welihinda, Hayes, Radford, et~al.]{hurst2024gpt}
Aaron Hurst, Adam Lerer, Adam~P Goucher, Adam Perelman, Aditya Ramesh, Aidan Clark, AJ~Ostrow, Akila Welihinda, Alan Hayes, Alec Radford, et~al.
\newblock Gpt-4o system card.
\newblock \emph{arXiv preprint arXiv:2410.21276}, 2024.

\bibitem[Jiang et~al.(2024)Jiang, Li, and Jamnik]{jiang2024multilingual}
Albert~Qiaochu Jiang, Wenda Li, and Mateja Jamnik.
\newblock Multilingual mathematical autoformalization, 2024.
\newblock URL \url{https://openreview.net/forum?id=QqdloE1QH2}.

\bibitem[LI et~al.(2024)LI, Beeching, Tunstall, Lipkin, Soletskyi, Huang, Rasul, Yu, Jiang, Shen, Qin, Dong, Zhou, Fleureau, Lample, and Polu]{numina_math_datasets}
Jia LI, Edward Beeching, Lewis Tunstall, Ben Lipkin, Roman Soletskyi, Shengyi~Costa Huang, Kashif Rasul, Longhui Yu, Albert Jiang, Ziju Shen, Zihan Qin, Bin Dong, Li~Zhou, Yann Fleureau, Guillaume Lample, and Stanislas Polu.
\newblock Numinamath.
\newblock \url{[https://huggingface.co/AI-MO/NuminaMath-CoT](https://github.com/project-numina/aimo-progress-prize/blob/main/report/numina_dataset.pdf)}, 2024.

\bibitem[Lin et~al.(2024)Lin, Sun, Welleck, and Yang]{lin24lean}
Haohan Lin, Zhiqing Sun, Sean Welleck, and Yiming Yang.
\newblock Lean-star: Learning to interleave thinking and proving.
\newblock In \emph{The 4th Workshop on Mathematical Reasoning and AI at NeurIPS'24}, 2024.

\bibitem[Luo et~al.(2023)Luo, Sun, Xu, Zhao, Lou, Tao, Geng, Lin, Chen, and Zhang]{luo2023wizardmath}
Haipeng Luo, Qingfeng Sun, Can Xu, Pu~Zhao, Jianguang Lou, Chongyang Tao, Xiubo Geng, Qingwei Lin, Shifeng Chen, and Dongmei Zhang.
\newblock Wizardmath: Empowering mathematical reasoning for large language models via reinforced evol-instruct.
\newblock \emph{arXiv preprint arXiv:2308.09583}, 2023.

\bibitem[Moura \& Ullrich(2021)Moura and Ullrich]{moura2021lean}
Leonardo~de Moura and Sebastian Ullrich.
\newblock The lean 4 theorem prover and programming language.
\newblock In \emph{Automated Deduction--CADE 28: 28th International Conference on Automated Deduction, Virtual Event, July 12--15, 2021, Proceedings 28}, pp.\  625--635. Springer, 2021.

\bibitem[PAULSON(1994)]{paulson1994isabelle}
LC~PAULSON.
\newblock Isabelle: A generic theorem prover.
\newblock \emph{Lecture Notes in Computer Science}, 828, 1994.

\bibitem[Tian et~al.(2024)Tian, Peng, Song, Jin, Yu, Mi, and Yu]{tian2024toward}
Ye~Tian, Baolin Peng, Linfeng Song, Lifeng Jin, Dian Yu, Haitao Mi, and Dong Yu.
\newblock Toward self-improvement of llms via imagination, searching, and criticizing.
\newblock In \emph{Proceedings of Neurips 2024}, 2024.

\bibitem[Wang et~al.(2024{\natexlab{a}})Wang, Song, Tian, Peng, Yu, Mi, Su, and Yu]{wang2024litesearch}
Ante Wang, Linfeng Song, Ye~Tian, Baolin Peng, Dian Yu, Haitao Mi, Jinsong Su, and Dong Yu.
\newblock Litesearch: Efficacious tree search for llm.
\newblock \emph{arXiv preprint arXiv:2407.00320}, 2024{\natexlab{a}}.

\bibitem[Wang et~al.(2024{\natexlab{b}})Wang, Deng, Lyu, Zeng, He, Yan, and An]{wang2024q}
Chaojie Wang, Yanchen Deng, Zhiyi Lyu, Liang Zeng, Jujie He, Shuicheng Yan, and Bo~An.
\newblock Q*: Improving multi-step reasoning for llms with deliberative planning.
\newblock \emph{arXiv preprint arXiv:2406.14283}, 2024{\natexlab{b}}.

\bibitem[Wang et~al.(2024{\natexlab{c}})Wang, Li, Shao, Xu, Dai, Li, Chen, Wu, and Sui]{wang2024math}
Peiyi Wang, Lei Li, Zhihong Shao, Runxin Xu, Damai Dai, Yifei Li, Deli Chen, Yu~Wu, and Zhifang Sui.
\newblock Math-shepherd: Verify and reinforce llms step-by-step without human annotations.
\newblock In \emph{Proceedings of the 62nd Annual Meeting of the Association for Computational Linguistics (Volume 1: Long Papers)}, pp.\  9426--9439, 2024{\natexlab{c}}.

\bibitem[Wei et~al.(2024)Wei, Sun, and Wang]{wei2024proving}
Chenrui Wei, Mengzhou Sun, and Wei Wang.
\newblock Proving olympiad algebraic inequalities without human demonstrations, 2024.
\newblock URL \url{https://openreview.net/forum?id=8kFctyli9H}.

\bibitem[Wu et~al.(2024)Wu, Huang, Zhou, Ying, Wang, Lin, and Chen]{wu2024internlm2}
Zijian Wu, Suozhi Huang, Zhejian Zhou, Huaiyuan Ying, Jiayu Wang, Dahua Lin, and Kai Chen.
\newblock Internlm2. 5-stepprover: Advancing automated theorem proving via expert iteration on large-scale lean problems.
\newblock \emph{arXiv preprint arXiv:2410.15700}, 2024.

\bibitem[Xin et~al.(2024{\natexlab{a}})Xin, Guo, Shao, Ren, Zhu, Liu, Ruan, Li, and Liang]{xin2024deepseekv1}
Huajian Xin, Daya Guo, Zhihong Shao, Zhizhou Ren, Qihao Zhu, Bo~Liu, Chong Ruan, Wenda Li, and Xiaodan Liang.
\newblock Deepseek-prover: Advancing theorem proving in llms through large-scale synthetic data.
\newblock \emph{arXiv preprint arXiv:2405.14333}, 2024{\natexlab{a}}.

\bibitem[Xin et~al.(2024{\natexlab{b}})Xin, Ren, Song, Shao, Zhao, Wang, Liu, Zhang, Lu, Du, Gao, Zhu, Yang, Gou, Wu, Luo, and Ruan]{xin2024deepseekproverv15}
Huajian Xin, Z.~Z. Ren, Junxiao Song, Zhihong Shao, Wanjia Zhao, Haocheng Wang, Bo~Liu, Liyue Zhang, Xuan Lu, Qiushi Du, Wenjun Gao, Qihao Zhu, Dejian Yang, Zhibin Gou, Z.~F. Wu, Fuli Luo, and Chong Ruan.
\newblock Deepseek-prover-v1.5: Harnessing proof assistant feedback for reinforcement learning and monte-carlo tree search, 2024{\natexlab{b}}.
\newblock URL \url{https://arxiv.org/abs/2408.08152}.

\bibitem[Xin et~al.(2024{\natexlab{c}})Xin, Ren, Song, Shao, Zhao, Wang, Liu, Zhang, Lu, Du, et~al.]{xin2024deepseek}
Huajian Xin, ZZ~Ren, Junxiao Song, Zhihong Shao, Wanjia Zhao, Haocheng Wang, Bo~Liu, Liyue Zhang, Xuan Lu, Qiushi Du, et~al.
\newblock Deepseek-prover-v1.5: Harnessing proof assistant feedback for reinforcement learning and monte-carlo tree search.
\newblock \emph{arXiv preprint arXiv:2408.08152}, 2024{\natexlab{c}}.

\bibitem[Yang et~al.(2024)Yang, Swope, Gu, Chalamala, Song, Yu, Godil, Prenger, and Anandkumar]{yang2024leandojo}
Kaiyu Yang, Aidan Swope, Alex Gu, Rahul Chalamala, Peiyang Song, Shixing Yu, Saad Godil, Ryan~J Prenger, and Animashree Anandkumar.
\newblock Leandojo: Theorem proving with retrieval-augmented language models.
\newblock \emph{Advances in Neural Information Processing Systems}, 36, 2024.

\bibitem[Ying et~al.(2024)Ying, Wu, Geng, Wang, Lin, and Chen]{ying2024lean}
Huaiyuan Ying, Zijian Wu, Yihan Geng, Jiayu Wang, Dahua Lin, and Kai Chen.
\newblock Lean workbook: A large-scale lean problem set formalized from natural language math problems.
\newblock \emph{arXiv preprint arXiv:2406.03847}, 2024.

\bibitem[Yu et~al.(2024)Yu, Jiang, Shi, Jincheng, Liu, Zhang, Kwok, Li, Weller, and Liu]{yumetamath}
Longhui Yu, Weisen Jiang, Han Shi, YU~Jincheng, Zhengying Liu, Yu~Zhang, James Kwok, Zhenguo Li, Adrian Weller, and Weiyang Liu.
\newblock Metamath: Bootstrap your own mathematical questions for large language models.
\newblock In \emph{The Twelfth International Conference on Learning Representations}, 2024.

\bibitem[Zheng et~al.(2022)Zheng, Han, and Polu]{zheng2022miniff}
Kunhao Zheng, Jesse~Michael Han, and Stanislas Polu.
\newblock minif2f: a cross-system benchmark for formal olympiad-level mathematics.
\newblock In \emph{International Conference on Learning Representations}, 2022.
\newblock URL \url{https://openreview.net/forum?id=9ZPegFuFTFv}.

\end{thebibliography}
\bibliographystyle{iclr2025_conference}

\newpage
\appendix
\def\lstlanguagefiles{lstlean.tex}
\section{Examples Theorems Proved by HunyuanProver}
\label{sec:examples}

\begin{tcolorbox}[
title=Example 1 (From Chinese High School Mathematics League),boxrule=0.5pt,colframe=purple!75!purple
]
\begin{tcolorbox}[boxrule=0pt]
\textcolor{blue}{Natural Language Problem} \\
The sequences $\{ a_n\}$ and $\{b_n\}$ satisfy $a_1=1, b_1=2, a_{n+1}=$
$\frac{1+a_n+a_nb_n}{b_n}, b_{n+1}=\frac{1+b_n+a_nb_n}{a_n}.$
Prove that $:a_{2008}<5.$
\end{tcolorbox}
\begin{lstlisting}[language=LEAN]
theorem theorem_11044972_a81f_4ea2_afa4_339d23e89245 (a b : ℕ → ℝ) 
(a1 : a 1 = 1) 
(b1 : b 1 = 2) 
(a_rec : ∀ n, a (n + 1) = (1 + a n + a n * b n) / b n) 
(b_rec : ∀ n, b (n + 1) = (1 + b n + a n * b n) / a n) : 
a 2008 < 5 := by

induction 2008 <;> simp_all [Nat.add_succ, Nat.succ_add, Nat.mul_succ, Nat.succ_mul]
contrapose! a1
all_goals have h1 : a 0 ≠ 0 := fun h => by simp [h] at b1
field_simp at b1 ⊢
intro h
all_goals
  have := a_rec 0
  have := b_rec 0
  simp_all [div_eq_mul_inv, mul_add, add_mul, mul_comm, mul_assoc, mul_left_comm]
field_simp at h
have h2 : b 0 ≠ 0 := fun hb => by simp [hb] at h
field_simp at h
nlinarith
have h2 : b 0 ≠ 0 := fun h => by simp [h] at a1
field_simp at a1 b1 ⊢
have h3 : 0 < 1 + a 0 + a 0 * b 0 := by nlinarith
have h4 := mul_self_pos.mpr h1
rw [div_lt_iff] <;> nlinarith
\end{lstlisting}
\label{tab:example1}
\end{tcolorbox}

\newpage

\begin{tcolorbox}[
title=Example 2 (From IMO\_1962\_P2),boxrule=0.5pt,colframe=purple!75!purple
]
\begin{tcolorbox}[boxrule=0pt]
\textcolor{blue}{Natural Language Problem} \\
Determine all real numbers $x$ which satisfy the inequality: \\
$$\sqrt{\sqrt{3-x}-\sqrt{x+1}}>\dfrac{1}{2}$$
Show that it is $\left[ ~ -1,\quad 1-\dfrac{\sqrt{127}}{32} ~ \right)$.
\end{tcolorbox}

\begin{lstlisting}[language=LEAN]
theorem imo_1962_p2 (x : ℝ) (h₀ : 0 ≤ 3 - x) (h₁ : 0 ≤ x + 1)
    (h₂ : 1 / 2 < Real.sqrt (Real.sqrt (3 - x) - Real.sqrt (x + 1))) : -1 ≤ x ∧ x < 1 - Real.sqrt 127 / 32 := by

rw [← sub_nonneg] at h₀ h₁
refine' ⟨by linarith [Real.sqrt_nonneg (3 - x), Real.sqrt_nonneg (x + 1)], _⟩
contrapose! h₂
norm_num at h₀ h₁
rw [Real.sqrt_le_iff]
norm_num
rw [Real.sqrt_le_left]
rw [sq, add_mul_self_eq]
field_simp
norm_num
rw [← sub_nonneg]
swap
refine' add_nonneg _ (sqrt_nonneg _)
linarith [Real.sqrt_nonneg (x + 1)]
have h₃ : 0 ≤ Real.sqrt (x + 1) := sqrt_nonneg _
nlinarith [Real.sq_sqrt h₁, Real.sq_sqrt (by positivity : (0:ℝ) ≤ 127)]
\end{lstlisting}
\label{tab:example2}
\end{tcolorbox}

\begin{tcolorbox}[
title=Example 3 (From lean\_workbook\_2061),boxrule=0.5pt,colframe=purple!75!purple
]
\begin{tcolorbox}[boxrule=0pt]
\textcolor{blue}{Natural Language Problem} \\
Prove the theorem: Let $a,b,c$ be three integer numbers so that $abc\ne 0$. Then the equation $ax+by=c$ has at least an integer solution if and only if the greatest common divisor of the numbers $a$ and $b$ divides the number $c$.
\end{tcolorbox}

\begin{lstlisting}[language=LEAN]
theorem lean_workbook_2061 (a b c : ℤ) (habc : a * b * c ≠ 0) :  (∃ x y : ℤ, a * x + b * y = c) ↔ (gcd a b) | c  := by

constructor
rintro ⟨x, y, h⟩
rw [← h]
exact dvd_add (dvd_mul_of_dvd_left (GCDMonoid.gcd_dvd_left a b) x)
    (dvd_mul_of_dvd_left (GCDMonoid.gcd_dvd_right a b) y)
intro h
obtain ⟨x, y, hxy⟩ := exists_gcd_eq_mul_add_mul a b
obtain ⟨z, rfl⟩ := h
exact ⟨x * z, y * z, by rw [← mul_assoc, ← mul_assoc, hxy]; ring⟩
\end{lstlisting}
\label{tab:example3}
\end{tcolorbox}

\begin{tcolorbox}[
title=Example 4 (From AIPS),boxrule=0.5pt,colframe=purple!75!purple
]
\begin{tcolorbox}[boxrule=0pt]
\textcolor{blue}{Natural Language Problem} \\
$abc/(a\sqrt(a^3 + c^3) + b\sqrt(a^3 + b^3) + c\sqrt(b^3 + c^3)) <= (a^2 + b^2)(a^2 + c^2)(b^2 + c^2)/(2((a^2 + b^2)(a^2 + c^2)\sqrt(a^3 + c^3) + (a^2 + b^2)\sqrt(a^3 + b^3)(b^2 + c^2) + (a^2 + c^2)(b^2 + c^2)\sqrt(b^3 + c^3)))$
\end{tcolorbox}

\begin{lstlisting}[language=LEAN]
theorem theorem_dc0af938_4a25_4093_8ed3_7f4999a556ed {a b c : ℝ} (ha : 0 < a) (hb : 0 < b) (hc : 0 < c) : a * b * c / (a * Real.sqrt (a ^ 3 + c ^ 3) + b * Real.sqrt (a ^ 3 + b ^ 3) + c * Real.sqrt (b ^ 3 + c ^ 3)) ≤ (a ^ 2 + b ^ 2) * (a ^ 2 + c ^ 2) * (b ^ 2 + c ^ 2) / (2 * (a ^ 2 + b ^ 2) * (a ^ 2 + c ^ 2) * Real.sqrt (a ^ 3 + c ^ 3) + 2 * (a ^ 2 + b ^ 2) * Real.sqrt (a ^ 3 + b ^ 3) * (b ^ 2 + c ^ 2) + 2 * (a ^ 2 + c ^ 2) * (b ^ 2 + c ^ 2) * Real.sqrt (b ^ 3 + c ^ 3)) := by

have h1 : 0 ≤ (a - b) ^ 2 * (a^2 + c^2) * (b^2 + c^2) := by positivity
rw [mul_assoc]
rw [← sub_nonneg]
have h2 : 0 ≤ (a - c) ^ 2 * (a ^ 2 + b ^ 2) * (b ^ 2 + c ^ 2) := by positivity
have h3 : 0 ≤ (b - c) ^ 2 * (a ^ 2 + b ^ 2) * (a ^ 2 + c ^ 2) := by positivity
field_simp [ha.ne', hb.ne', hc.ne', h1, h2, h3]
have h4 : 0 ≤ a ^ 2 * b ^ 2 := mul_nonneg (sq_nonneg _) (sq_nonneg _)
have h5 : 0 ≤ (a * Real.sqrt (a ^ 3 + c ^ 3) - b * Real.sqrt (a ^ 3 + b ^ 3)) ^ 2 := sq_nonneg _
have h6 := sq_nonneg (a * Real.sqrt (b^3 + c^3) - b * Real.sqrt (a^3 + c^3))
have h7 : 0 ≤ (a * Real.sqrt (a ^ 3 + c ^ 3) - c * Real.sqrt (b ^ 3 + c ^ 3)) ^ 2 := sq_nonneg _
have h8 : 0 ≤ (b * Real.sqrt (a ^ 3 + c ^ 3) - c * Real.sqrt (a ^ 3 + b ^ 3)) ^ 2 := sq_nonneg _
have h9 : 0 ≤ a ^ 2 * c ^ 2 := mul_nonneg (sq_nonneg _) (sq_nonneg _)
have h10 : 0 ≤ (b^2 + c^2) * (a^2 + c^2) := by positivity
have h11 : 0 ≤ (a^2 + b^2) * (a^2 + c^2) := by positivity
have h12 : 0 ≤ (a^2 + b^2) * (b^2 + c^2) := by positivity
have h13 : 0 ≤ (a * Real.sqrt (a ^ 3 + c ^ 3) - b * Real.sqrt (a ^ 3 + b ^ 3)) ^ 2 := sq_nonneg _
have h14 := sq_nonneg (a * Real.sqrt (b ^ 3 + c ^ 3) - c * Real.sqrt (a ^ 3 + c ^ 3))
have h15 : 0 ≤ (a * Real.sqrt (a^3 + c^3) - b * Real.sqrt (a^3 + b^3))^2 := sq_nonneg _
have h16 : 0 ≤ (a * Real.sqrt (a ^ 3 + b ^ 3) - b * Real.sqrt (b ^ 3 + c ^ 3)) ^ 2 := sq_nonneg _
have h17 := sq_nonneg (a * Real.sqrt (b^3 + c^3) - b * Real.sqrt (a^3 + c^3))
have h18 : 0 ≤ (a * Real.sqrt (a^3 + c^3) - b * Real.sqrt (a^3 + b^3))^2 := sq_nonneg _
have h19 := sq_nonneg (a * Real.sqrt (a^3 + c^3) - c * Real.sqrt (b^3 + c^3))
have h20 : 0 ≤ a ^ 2 * b ^ 2 := mul_nonneg (sq_nonneg _) (sq_nonneg _)
have h21 : 0 ≤ a * Real.sqrt (a^3 + c^3) := mul_nonneg ha.le (Real.sqrt_nonneg _)
have h22 : 0 ≤ b * Real.sqrt (a ^ 3 + b ^ 3) := by positivity
have h23 : 0 ≤ c * Real.sqrt (b ^ 3 + c ^ 3) := by positivity
refine' div_nonneg (by nlinarith) (by positivity)
\end{lstlisting}
\label{tab:example4}
\end{tcolorbox}

\begin{tcolorbox}[
title=Example 5 (From NuminaMath),boxrule=0.5pt,colframe=purple!75!purple
]
\begin{tcolorbox}[boxrule=0pt]
\textcolor{blue}{Natural Language Problem} \\
Given the function $f(x)= \begin{cases} 2^{1-x} & (x \leqslant 1) \\ 1 - \log_{2}x & (x > 1) \end{cases}$, find the range of $x$ that satisfies $f(x) \leqslant 2$. \\
\\
A: $[-1, 2]$ \\
B: $[0, 2]$ \\
C: $[1, +\infty)$ \\
D: $[0, +\infty)$ \\
\\
answer: $[0, +\infty)$ \\
\end{tcolorbox}

\begin{lstlisting}[language=LEAN]
theorem theorem_5319d37e_b470_42f5_a42f_b4221daceb81 (f : ℝ → ℝ) (x : ℝ) (hf: f x = if x ≤ 1 then 2^(1-x) else 1 - Real.logb 2 x) : f x ≤ 2 ↔ x ≥ 0 := by

rw [hf]
refine' ⟨fun h => _, fun h => _⟩
split_ifs at h with h_1
contrapose! h
rw [← Real.rpow_one 2]
rw [← rpow_mul (by norm_num)]
rw [one_mul, Real.rpow_lt_rpow_left_iff (by norm_num)]
linarith
all_goals aesop
case neg =>
  linarith
rw [← Real.rpow_one 2]
rw [← Real.rpow_mul zero_lt_two.le]
rw [one_mul, Real.rpow_le_rpow_left_iff (by norm_num)]
linarith
linarith [logb_pos (show (1 : ℝ) < 2 by norm_num) <_>]
\end{lstlisting}
\label{tab:example5}
\end{tcolorbox}

\section{Policy Prompt}
\label{sec:policy_prompt}

\begin{tcolorbox}[
colback=white!10!white,
colframe=purple!75!purple,
title=An example of the policy model
]
\textcolor{blue}{Prompt}
\begin{lstlisting}[language=LEAN]
Given the Lean 4 tactic state, suggest a next tactic.
Tactic state:
a b : ℝ
hab : a > 0 ∧ b > 0
h1 : 0 < Real.log 2
h : Real.log (a * b) = Real.log (2 ^ 6)
⊢ a + b ≥ 16

Next tactic:
\end{lstlisting}
\textcolor{blue}{Response}
\begin{lstlisting}[language=LEAN]
rw [Real.log_mul] at h
\end{lstlisting}
\label{tab:dataexample1}
\end{tcolorbox}

\section{Distance Critic Prompt}
\label{sec:distance_critic_prompt}

\begin{tcolorbox}[
colback=white!10!white,
colframe=purple!75!purple,
title=An example of the critic model
]
\textcolor{blue}{Prompt}
\begin{lstlisting}[language=LEAN]
Here is a Lean4 state:
case refine'_1.intro
a : ℝ
h : (∀ x ∈ Set.Icc 0 1, a ≥ rexp x) ∧ ∃ x, x ^ 2 + 4 * x + a = 0
hx1 : ℝ
hx2 : hx1 ^ 2 + 4 * hx1 + a = 0
⊢ a ∈ Set.Icc (rexp 1) 4

case refine'_2
a : ℝ
h : a ∈ Set.Icc (rexp 1) 4
x : ℝ
hx : x ∈ Set.Icc 0 1
⊢ a ≥ rexp x

case refine'_3
a : ℝ
h : a ∈ Set.Icc (rexp 1) 4
⊢ ∃ x, x ^ 2 + 4 * x + a = 0

----------------
Please tell me how many more tactic steps are needed to finish this state.

\end{lstlisting}
\textcolor{blue}{Response}
\begin{lstlisting}[language=LEAN,mathescape=true]
Let me think step by step.
... ...
So, there is <num_box><num-1-of-2><num-1-of-4><num-1-of-8>
<num-2-of-16><num-4-of-32><num-7-of-64></num_box> more tactic steps are needed to finish this state.
\end{lstlisting}
\label{tab:dataexample2}
\end{tcolorbox}

\end{document}